%% file: main.tex
\documentclass[conference]{IEEEtran}
\IEEEoverridecommandlockouts
\usepackage{cite}
\usepackage{amsmath,amssymb,amsfonts}
\usepackage[ruled,vlined,linesnumbered]{algorithm2e}
\usepackage{algpseudocode}
\usepackage[hidelinks]{hyperref}
\usepackage{csquotes}
\usepackage[inline]{enumitem}
\usepackage{graphicx}
\usepackage{textcomp}
\usepackage{xcolor}
\def\BibTeX{{\rm B\kern-.05em{\sc i\kern-.025em b}\kern-.08em
    T\kern-.1667em\lower.7ex\hbox{E}\kern-.125emX}}



\input{misc}
    
\begin{document}

\title{
    Adversarial Predictions of Data Distributions Across Federated 
    Internet-of-Things Devices
}

\author{
    \IEEEauthorblockN{
        Samir Rajani\IEEEauthorrefmark{1}, 
        Dario Dematties\IEEEauthorrefmark{1}\IEEEauthorrefmark{2}, 
        Nathaniel Hudson\IEEEauthorrefmark{3}\IEEEauthorrefmark{4}, 
        Kyle Chard\IEEEauthorrefmark{3}\IEEEauthorrefmark{4}, 
        Nicola Ferrier\IEEEauthorrefmark{1}\IEEEauthorrefmark{2}, \\ 
        Rajesh Sankaran\IEEEauthorrefmark{1}\IEEEauthorrefmark{2}, 
        and 
        Peter Beckman\IEEEauthorrefmark{1}\IEEEauthorrefmark{2}
    }
    \IEEEauthorblockA{
        \IEEEauthorrefmark{1}%
        Northwestern Argonne Institute of Science and Engineering, Northwestern University; Evanston, IL, United States
    }
    \IEEEauthorblockA{
        \IEEEauthorrefmark{2}%
        Mathematics and Computer Science Division, Argonne National Laboratory; Lemont, IL, United States
    }
    \IEEEauthorblockA{
        \IEEEauthorrefmark{3}%
        Department of Computer Science, University of Chicago; Chicago, IL, United States
    }
    \IEEEauthorblockA{
        \IEEEauthorrefmark{4}%
        Data Science and Learning Division, Argonne National Laboratory; Lemont, IL, United States
    }
}

\maketitle

\begin{abstract}
    Federated learning~(FL) is increasingly becoming the default approach for training machine learning models across decentralized Internet-of-Things~(IoT) devices. A key advantage of FL is that no raw data are communicated across the network, providing an immediate layer of privacy. Despite this, recent works have demonstrated that data reconstruction can be done with the locally trained model updates which are communicated across the network. However, many of these works have limitations with regard to how the gradients are computed in backpropagation. In this work, we demonstrate that the model weights shared in FL can expose revealing information about the local data distributions of IoT devices. This leakage could expose sensitive information to malicious actors in a distributed system. We further discuss results which show that injecting noise into model weights is ineffective at preventing data leakage without seriously harming the global model accuracy.
\end{abstract}


\section{Introduction}
\label{sec:introduction}

\input{sections/1_intro}

\section{Related Work}
\label{sec:related_work}
\input{sections/2_related_work_v2}

\section{Problem Description \& Research Questions}
\label{sec:problem}
\input{sections/4_questions}

\section{Experimental Design \& Results}
\label{sec:experiments}
\input{sections/5_results}

\section{Conclusions \& Future Directions}
\label{sec:conclusion}
\input{sections/6_conclusion}

\section*{Acknowledgment}
We want to thank Bhupendra A. Raut, Seongha Park, Robert C. Jackson, Sean Shahkarami, Yongho Kim, and Ian Foster for their insightful comments, ideas, and support. This work was supported in part by U.S. Department of Energy, Office of Science, under contract  DE-AC02-06CHI1357, National Science Foundation's Mid-Scale Research Infrastructure grant, NSF-OAC-1935984, and U.S. Department of Energy, Office of Science, under grant DOE-145-SE-PRJ1009506. This research used resources of the Argonne Leadership Computing Facility, which is a DOE Office of Science User Facility supported under Contract DE-AC02-06CH11357.

\bibliographystyle{ieeetr}
\bibliography{
}


\end{document}

%% file: misc.tex
\usepackage{todonotes}  

%% file: sections/1_intro.tex
\begin{figure}
	\centering
    \includegraphics[width=0.8\linewidth]{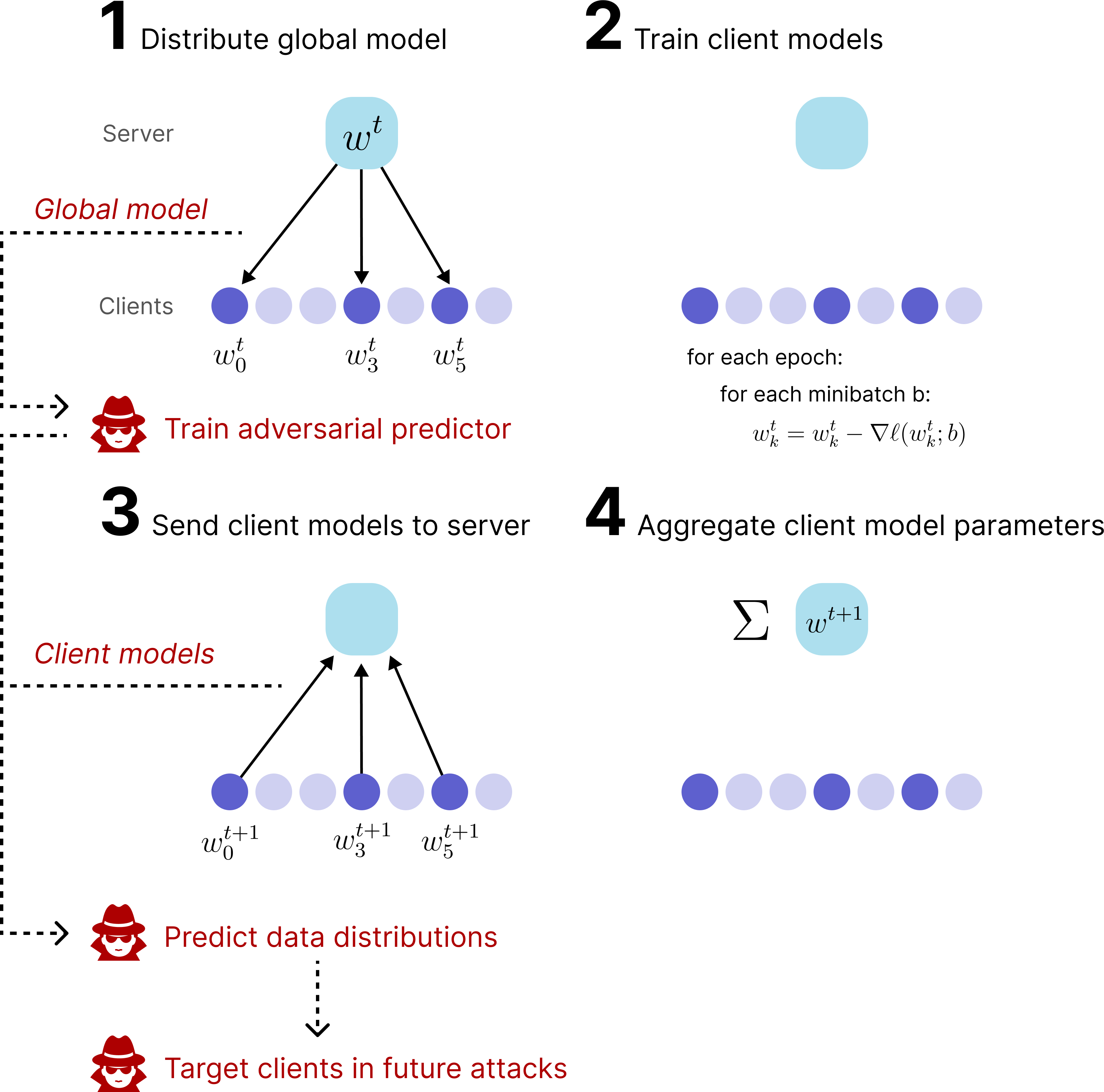}
    \caption{The standard steps for FL process: global model distribution, client model training, client model communication, and aggregation. The steps in red outline the procedure by which an adversary might conduct an attack.}
    \label{fig:system_model}
\end{figure}

\textit{Federated Learning}~(FL) is a method for decentralized machine learning which can seamlessly train models across many \textit{Internet-of-Things}~(IoT) devices. In conventional decentralized machine learning, IoT devices send their training data to a central location (e.g., remote cloud server) to train a deep neural network or some other model. This approach is unattractive due to its high communication cost~\cite{hudson2021framework} and its requirement that all data---including possibly sensitive data---be shared across a network~\cite{Rieke_2020}. Under FL, no raw data are communicated over the network. Instead, the central server is sent only locally trained model weights provided by the IoT devices. In short, FL begins with a central server initializing a \enquote{global} machine learning model with random weights, and then the FL training loop begins. The FL training loop, depicted in \autoref{fig:system_model}, is as follows:
\begin{enumerate*}[label=\textit{(\roman*)}]
    \item the central server shares the current global model with the IoT devices;
    \item the IoT devices receive a copy of the global model and train it on their private data;
    \item the IoT devices send back their locally trained copies of the model to the central server;
    \item the central server aggregates all the received locally trained models to update the global model;
    and the loop repeats.
\end{enumerate*}

Because no raw data are communicated over the network in FL, it is a promising method for training a model on sensitive or private data. However, since FL requires that clients send their model updates across a network, the extent to which FL actually protects client data is unclear. 

Client models are trained on local data, meaning they could contain implicit revealing information about private training samples. Recent work has shown that gradients from a small batch of samples can be used to produce pixel-wise reconstructions of images and token-wise reconstructions of text~\cite{zhu2019deep}. While these findings are alarming, their application is limited to settings in which the gradients being transmitted are from a single, small batch of at most eight samples~\cite{zhu2019deep}, or to a limited extent when gradients are averaged over a larger number of samples~\cite{geiping2020inverting}. Other work~\cite{zhao2020idlg} has shown that it is possible to analytically reconstruct ground-truth labels for training samples from gradients, but this technique is viable only when gradients are known for every single training sample. By contrast, in typical FL settings, client models undergo one or more epochs of training over their local data before their parameters are sent to the server, rather than just a single gradient update on a small batch of samples. 

In this work, we show that even in realistic federated settings, sharing client model gradients or parameters leaks private information about training samples. In particular, we show that an adversary could train a deep neural network to extract a client's label distribution from its shared model parameters with near-perfect accuracy. In many federated settings, the ability of an adversary to predict data distributions of IoT devices can be dangerous. Consider, for example, a federated language modelling task in which the words of text messages sent by users of mobile phones serve as labels. In this setting, it is conceivable that an adversary who is able to reconstruct the labels of the samples on client devices might be able to deduce identifiable information about a user. An adversary might, for example, be able to make a list of clients whose text messages contain information related to banking, and then target these clients in the future, with more refined attacks. Moreover, the clients participating in FL might be under the false assumption that their privacy is assured because training samples remain on their device. To understand the extent to which user data is protected, we need a deeper understanding of what kinds of attacks are possible and what kinds of mitigation strategies are effective.

Our contributions can be summarized as follows: 
1) we show empirically that client model parameters contain information about the labels of samples on which they were trained; 
2) we perform several experiments that suggest explanations for the presence of this information; 3) we demonstrate that a deep neural network can be trained to recover information about the distributions of client labels; 
and 
4) we test the ability of simple defenses, such as adding Gaussian and Laplacian noise to gradients, to prevent the leakage of information about client label distributions. 
Our findings have significant implications for the use of FL in privacy-sensitive settings.

%% file: sections/2_related_work_v2.tex
Much of the popularity of FL can be accredited to the intuitive idea that it improves data privacy in decentralized machine learning since no raw data are communicated across the network~\cite{mcmahan2017communication, bonawitz2019towards, yang2018applied}. However, recent works have begun to challenge this assumption~\cite{zhao2020idlg}. Zhu et al. coin the notion of \enquote{deep leakage from gradients} to refer their demonstrated optimization-based approach to reconstruct data from gradient updates~\cite{zhu2019deep}. Similarly, Geiping et al. demonstrate a numerical reconstruction method which reconstructs images from a neural network gradient update~\cite{geiping2020inverting}.  However, these works have to relax the problem of data reconstruction substantially to demonstrate the effectiveness of their techniques. Specifically, their approaches are most effective when the gradient update from training the model is done on a single image. They both note that multi-image reconstruction can be done but that effectiveness of the technique suffers when gradients are averaged across modest batch sizes $B \geq 8$.

These growing concerns around data leakage from model updates have encouraged exploration into additional privacy-preserving techniques for FL. Some of the most popular include homomorphic encryption~\cite{homomorphic_encr_survey,wibawa2022homomorphic} and differential privacy~\cite{dwork_calibrating_2006, adnan_federated_2022, ji2014differential, andrew2023oneshot}. Homomorphic encryption is a technique that encrypts data in such a way that they are equivariant to many mathematical transformations. This means that the global model can be updated by an average of encrypted model weights. These averaged model weights can then be decrypted to train the models locally without corrupting the local models~\cite{9682053}. A central concern surrounding homomorphic encryption for FL is that the resource cost of encrypting data on resource-constrained IoT devices is non-trivial~\cite{jin2023fedmlhe,254465}. Differential privacy takes a different, less resource-costly approach. This technique directly injects an amount of noise, $\epsilon>0$, into client model updates before they are sent to aggregation node~\cite{wei2020federated, el2022differential}. While cheaper to perform, there is an obvious trade-off to consider. If $\epsilon$ is too large, the global model's accuracy will suffer; if $\epsilon$ is too small, privacy benefits are reduced.

Our work takes a fundamentally different approach from prior works in the FL privacy literature. While the works on data reconstruction from model gradients are critically important in the FL domain, we argue that they rely on unrealistic simplifying assumptions related to the amount of data model updates are trained on to support the reconstruction. Instead, we consider a more realistic scenario where the amount of data on IoT devices is larger than a handful of data samples (i.e., $\gg 8$). We then show that even with this more challenging scenario, the local data \textit{distributions} of IoT devices in an FL system can still be approximated based on the model weights.

%% file: sections/4_questions.tex
We consider a standard two-tier FL system comprised of decentralized IoT devices with local, private data and a central aggregation node which manages the FL process. We seek to learn the extent to which client information is compromised by sharing client gradients. In particular, could an adversary use the model parameters shared by clients to deduce information about their private training data?

Further, we study an adversary who has access to information about the federated training (e.g., an honest-but-curious server). In the case of a supervised objective, we assume the adversary has access to the dataset labels and their order in the output layer of the network. In the case of an unsupervised objective, as we will show in Section \ref{vcmp}, models representations contain relevant information about the input features of the data, and an adversary could potentially still deduce information about client data distributions. We also assume the adversary has access to the client optimization parameters, such as the learning rate $\eta$ and the number of local training epochs $E$. We focus on the possibility that an adversary could deduce information about the distribution of a client's training samples. Client training data are often not independently and identically distributed (non-IID), and their data distributions may contain private information. 

Specifically, we are interested in answering the following questions:
\begin{enumerate*}[label=\textbf{Q\arabic*.}]
    \item What is the relationship between client model parameters and the data distribution on which the parameters were learned?
    \item Can a client model's parameters be used directly to predict the distribution on which they were learned?
\end{enumerate*}
We expand on our approach to each of these questions below.

\subsection{Visualizing Client Model Parameters}

To understand the relationship between client model parameters and the data distribution on which they were learned, we train clients on a variety of data distributions and visualize their model parameters using principal component analysis (PCA). We refer to clients trained on synthetic data distributions as ``dummy clients," and to the space of dimension-reduced model parameters as a ``model-latent space."

In \autoref{sec:experiments}, we show several experiments which explain the relationship between the position of a client model in the model-latent space to the client's data distribution. By performing a layer-by-layer decomposition of the client model parameters, and by examining the behavior of client model parameters when the training objective is unsupervised, we also propose an explanation for this clustering.

\subsection{Predicting Client Data Distributions}

The existence of a relationship between client model parameters and the data distribution used for learning them raises a second question: can a model's position in the model-latent space be used directly to predict a client's label distribution? To answer this question, we introduce the notion of a ``meta-dataset." For client models trained on disjoint training samples $\mathcal{P}_k$ from some original dataset, the meta-dataset is a mapping from client model parameters $w_k$ to label distributions $\text{Pr}(y = y_i \text{ } \vert \text{ } i \in \mathcal{P}_k)$. We propose that an adversary might be capable of training a deep neural network on a synthetic meta-dataset, enabling them to extract information about clients.

In Algorithm~\ref{alg:adversary}, we propose one potential strategy an adversary might use to obtain information about client data distributions. Later, in \autoref{sec:experiments}, we validate empirically the effectiveness of this strategy in the presence of a simple defense: the addition of Gaussian and Laplacian noise to gradients. The adversary first intercepts the global model $w_t$ sent by the server to the clients (line 1). They then construct a meta-dataset by creating a large number of dummy clients whose local training data comes from a synthetic non-IID distribution (lines 3-6). The client models are initialized with the intercepted global model parameters and trained on their local training data. Then, their trained parameters $\hat{w}^k$ are reduced to some small number of dimensions using PCA (lines 8-10). Each dummy client provides one training sample in the meta-dataset, in which the PCA-reduced model parameters are the inputs, and the label distributions are the outputs. The adversary then trains a neural network on the meta-dataset (lines 11-14); the dimension of the PCA-reduced model parameters $\hat{w}^k_\text{pca}$ should be sufficiently small such that they can fit in the input layer. A key assumption is that, in order to train the ``dummy" client models, the adversary can access a set of proxy training samples that are similar to the training samples on the client devices. Upon intercepting model parameters sent from the clients to the server, they could apply the same PCA model to these parameters, and use their trained label distribution predictor to perform inference (lines 15-17).

\input{pseudocode/adversarial_pred}

%% file: pseudocode/adversarial_pred.tex
\begin{algorithm}[t]
\label{alg:adversary}
\DontPrintSemicolon
\caption{Adversarial data distribution prediction}
intercept global model weights, $w_t$\;
$\hat{S}^t \gets$ set of dummy clients\;
\ForEach{dummy client $k \in \hat{S}^t$}{
$\hat{\mathcal{P}}^k \gets \text{(random subset of proxy data)}$\;
$\hat{w}^k \gets \mathit{ClientUpdate}(k, w_t)$\;
$y^k \gets \text{Pr}(y = y_i \text{ } \vert \text{ } i \in \mathcal{P}_k)$\;
}

$W_\text{pca} \gets \text{(fit PCA model on the $\hat{w}^k$)}$\;

\ForEach{dummy client $k \in \hat{S}^t$}{
    $\hat{w}^k_\text{pca} \gets \text{(apply $W_{\text{pca}}$ to $\hat{w}^k$)}$
}

initialize label distribution predictor $f_\theta$\;
$\mathcal{B} \gets$ (split the $(\hat{w}^k_\text{pca}, y^k)$ into batches of size $\hat{B}$)\;
\ForEach{local epoch $e = 1, \ldots, \hat{E}$} {
    \ForEach{batch $b \in \mathcal{B}$} {
        $\theta \gets \theta - \eta'\nabla\hat{\ell}(\theta; b)$\;
    }
}
\ForEach{client $k \in S_t$} {
    intercept client model weights $w_{t + 1}^k$\;
    $y^k_\text{(pred)} \gets f_\theta(w_{t + 1}^k)$\;
}

\setcounter{AlgoLine}{0}
\SetKwProg{myproc}{Procedure}{}{}
\myproc{ClientUpdate$(k,w)$}{
    $\mathcal{B} \gets \text{(split $\hat{\mathcal{P}}_k$ into batches of size $B$)}$\;
    \ForEach{local epoch $e=1, \ldots, E$}{
        \ForEach{batch $b \in \mathcal{B}$}{
            $w \gets w - \eta\nabla\ell(w; b)$\;
        }
    }
    \Return $w$ to server\;
}
\end{algorithm}

%% file: sections/5_results.tex
\subsection{Visualizing Client Model Parameters}
\label{vcmp}

As a starting point for understanding the relationship between client model parameters and their label distributions, we replicate and extend an experiment performed by Wang et al.~\cite{wang2020optimizing}. We initialize a global model, which is broadcast to 100 clients with disjoint training sets. For each of the ten labels in the dataset, ten clients are provided with a synthetic non-IID training set, according to one of two sampling schemes. In the \enquote{80\%-20\% sampling} scheme, we provide each client with a data distribution in which 80\% of the samples have one label, and the remaining 20\% are uniformly distributed across the remaining labels. In the \enquote{Dirichlet sampling} scheme, client data distributions are sampled from a Dirichlet distribution. Each client is trained for one epoch, and all client model parameters are reduced to two dimensions via PCA for visualization.

\begin{figure}
	\centering
    \includegraphics[width=\columnwidth]{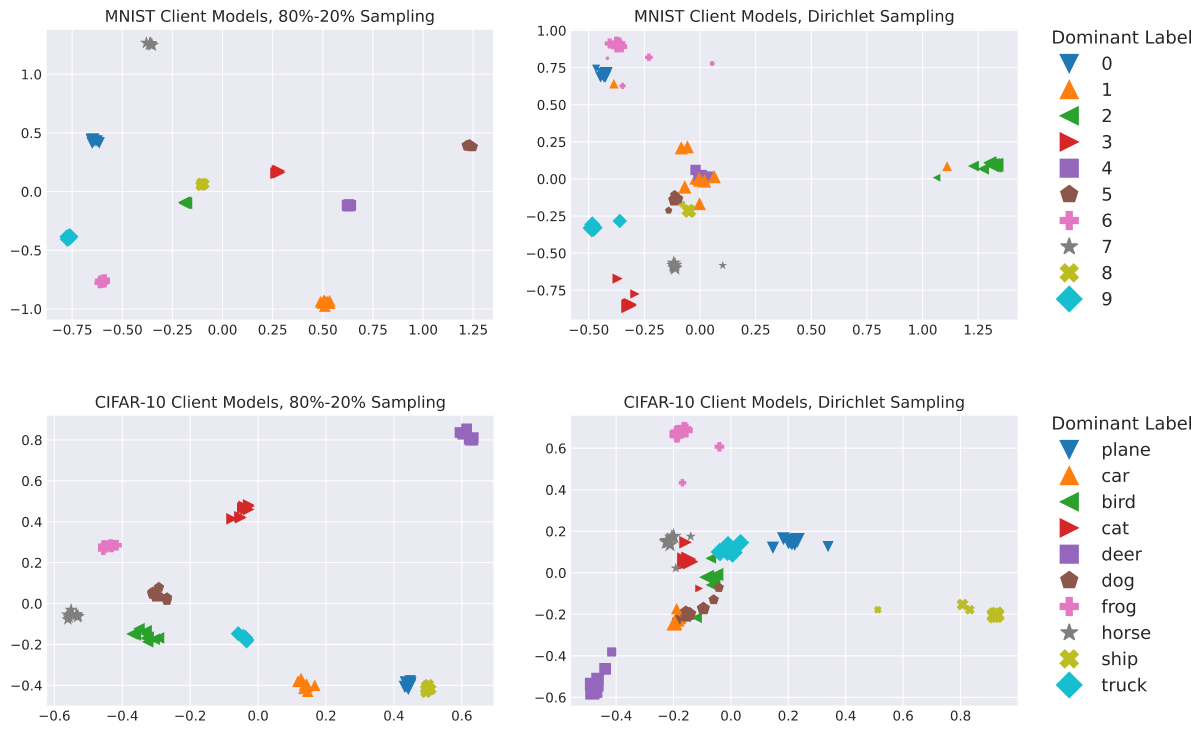}
    \caption{Clustering of PCA-reduced client model parameters on the MNIST and CIFAR-10 datasets according to the dominant label in their training sets. In the ``80\%-20\% sampling" scheme, the dominant label accounted for 80\% of a client's training samples. In the ``Dirichlet sampling" scheme, client data distributions were sampled from a Dirichlet distribution with $\alpha = 0.2$; smaller points denote that a smaller percentage of samples come from the dominant class, and these points appear to drift from the centers of their clusters. These experiments used a small learning rate ($\eta = 10^{-5}$), which we found to show the clearest clustering, but the same phenomenon was observed using all learning rates we tested.}
    \label{fig:dominant_label}
\end{figure}

We perform experiments using supervised classification tasks on the MNIST dataset using a multi-layer perceptron (MLP), and on the CIFAR-10 dataset using a convolutional neural network (CNN); the results of these experiments are shown in \autoref{fig:dominant_label}. For both datasets and architectures, we observe the same phenomenon: client model parameters are clustered according to their dominant labels, suggesting that they contain implicit information about the samples on which they were learned. We might hypothesize that this clustering originates from one or both of two distinct phenomena:

\begin{enumerate}
    \item Images belonging to the same class have similar features, leading clients with the same dominant label to train models with similar parameters.
    \item Images belonging to the same class have the same labels, and the nature of the supervised classification task means client models trained on sample distributions with the same dominant label will have similar parameters. Zhao et al. demonstrate a related phenomenon analytically for gradients from a single training sample~\cite{zhao2020idlg}.
\end{enumerate}

If the clustering is a result of the first phenomenon, we would expect that client models with semantically similar dominant labels (e.g., car and truck) would be close together in the model-latent space, and that clustering would occur even with an unsupervised training objective. Meanwhile, if the clustering is a result of the second phenomenon, we would expect the positions of dominant-label clusters to be arbitrary.

\begin{figure}
	\centering
    \includegraphics[width=0.85\linewidth]{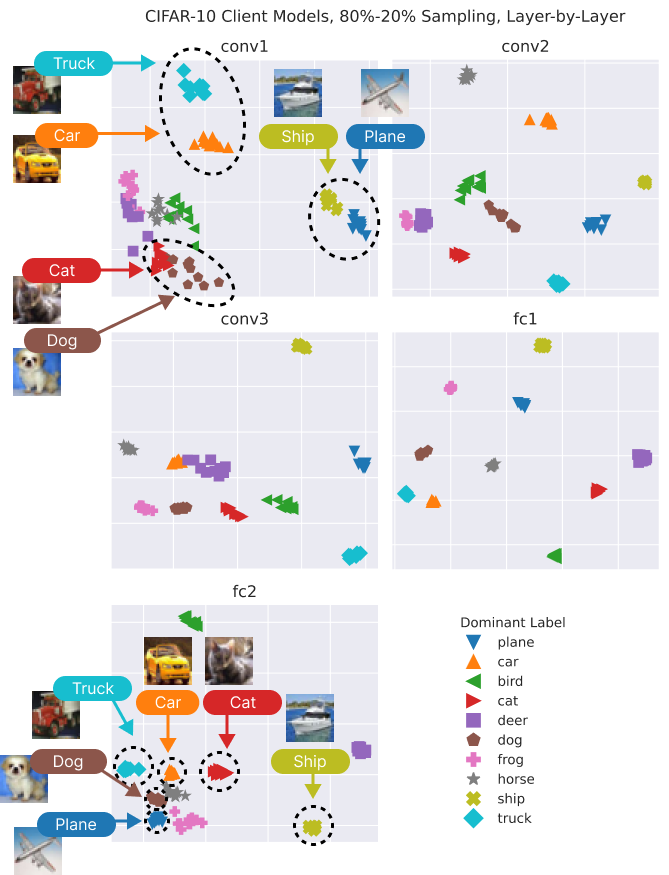}
    \caption{Layer-by-layer clustering of PCA-reduced client model parameters on the CIFAR-10 dataset. Semantic clustering appears to occur in the earlier convolutional layers, suggesting that the clustering is the result of the similarity in features of images belonging to the same class. Note, for example, that client models trained primarily on images of animals are concentrated in the bottom-left corner of the model-latent space for the first convolutional layer. Meanwhile, the clustering appears roughly arbitrary in the final fully-connected layer, suggesting that the clustering here is a result of labels.}
    \label{fig:layer_by_layer}
\end{figure}

To understand which of these phenomena accounts for the clustering, we perform two more experiments. First, we examine how the model-latent space behaves when we study each layer of the neural network in isolation. \autoref{fig:layer_by_layer} shows the PCA-reduced client model parameters for each layer of the CNN used for CIFAR-10. We find in the earlier convolutional layers, client models trained on distributions with semantically similar dominant labels (e.g., cat and dog, car and truck, plane and ship) are close together in the model-latent space. In the final fully-connected layers, this semantic clustering disappears, suggesting that the clustering is primarily a result of the labels provided in the supervised classification task. We hypothesize, therefore, that the formation of dominant-label clusters is a result of both phenomena we identified, the first manifesting in the earlier layers, which extract features from input samples, and the second manifesting in the later layers, designed for the classification task.

\begin{figure}
	\centering
    \includegraphics[width=\columnwidth]{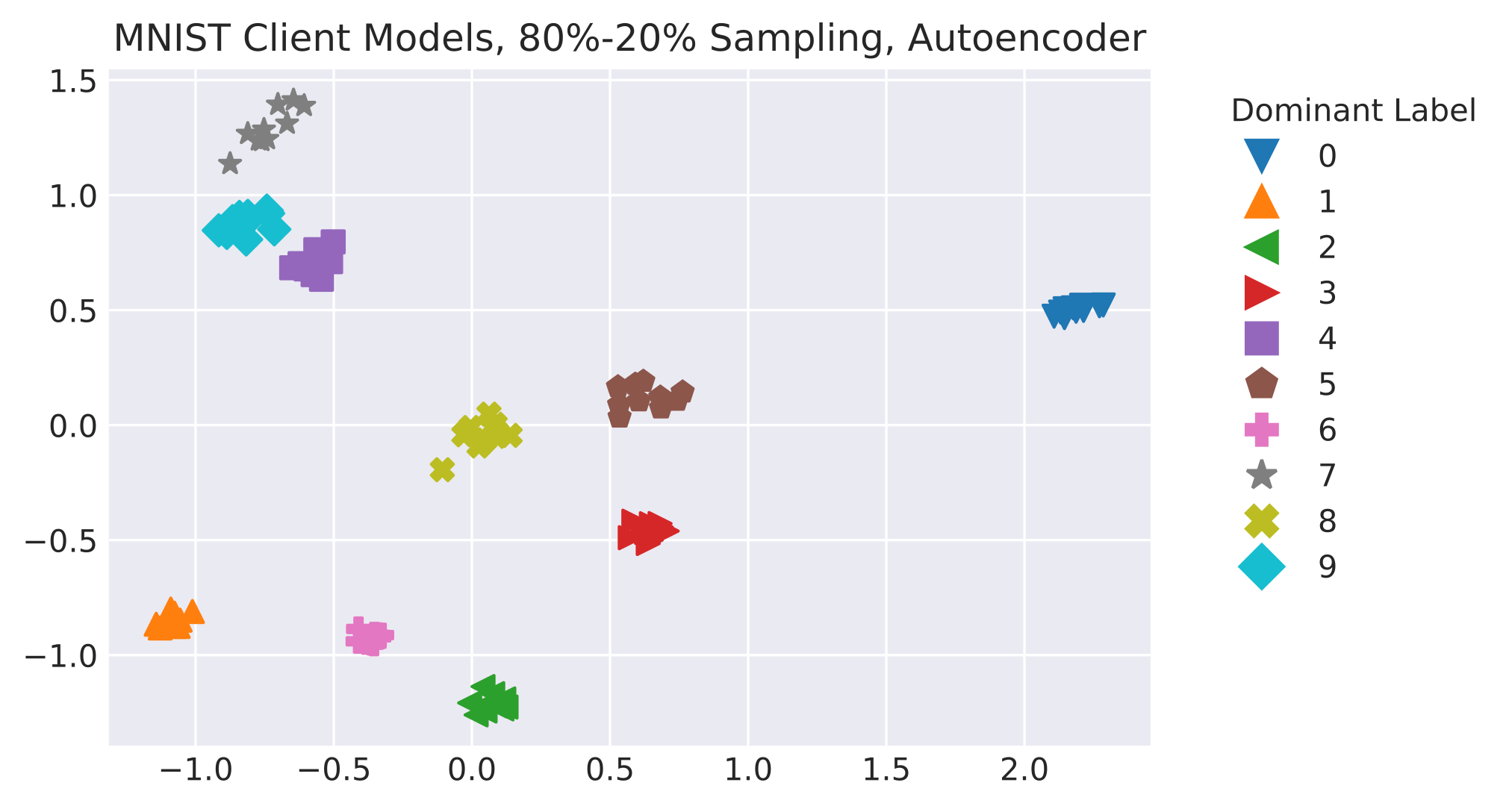}
    \caption{Clustering of autoencoder parameters trained with an unsupervised objective on MNIST. The clustering phenomenon occurs even in the absence of labels, suggesting that on simple datasets, it can be attributed in part to the similarity of features in training samples with the same label.}
    \label{fig:mnist_autoencoder}
\end{figure}

To validate this hypothesis, we perform a second experiment, in which we train an autoencoder on the MNIST dataset. If part of the clustering phenomenon can be accounted for by feature similarity, we would expect some clustering to occur even with an unsupervised task. Indeed, as shown in \autoref{fig:mnist_autoencoder}, we observe a strong clustering effect in the model-latent space according to the dominant label. The problem of label deficiency at the edge suggests the application of FL in unsupervised and self-supervised settings~\cite{he2021ssfl}. The finding that clustering occurs even when labels are not used in training suggests that in these applications, and in instances where the adversary does not have access to the labels used for a supervised objective, client privacy might be leaked.

\subsection{Predicting Client Data Distributions}

To determine the extent to which the adversarial strategy proposed in Algorithm \ref{alg:adversary} might work in a practical federated setting, and to test the viability of adding noise to gradients as a defense against this strategy, we run experiments on the MNIST and CIFAR-10 datasets, injecting three different scales of Gaussian and Laplacian noise into client gradients.

\begin{figure}
	\centering
    \includegraphics[width=0.85\linewidth]{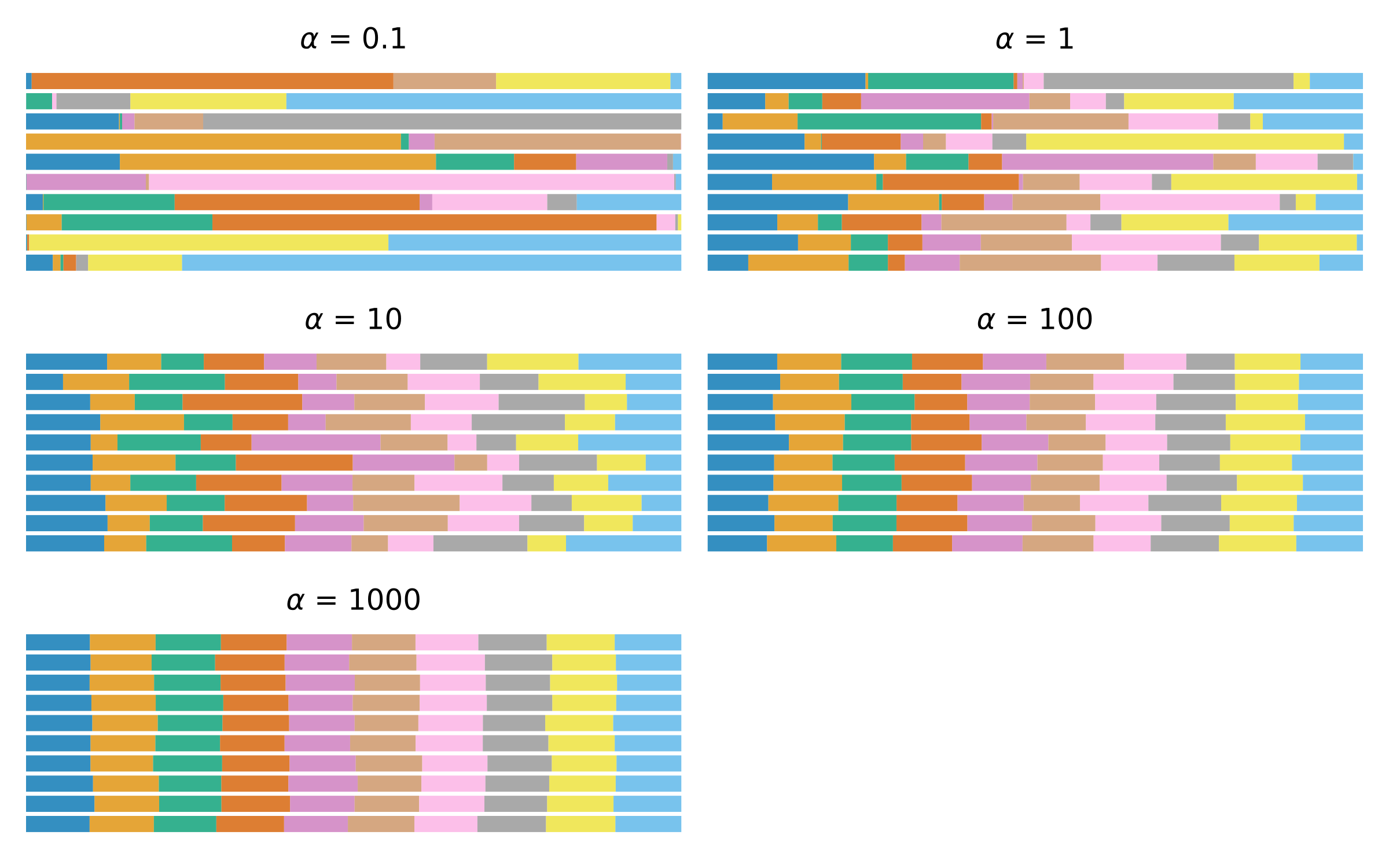}
    \caption{Example client label distributions sampled from a Dirichlet distribution, used to create synthetic non-IID data. We constructed a meta-dataset by training clients which had sample distributions drawn using concentration parameters $\alpha \in \{0.1, 1, 10, 100, 1000\}$, which provided a wide range of uniformities.}
    \label{fig:dirichlet_examples}
\end{figure}

For our experiments, the training set of the meta-dataset contains client model parameters trained on samples from the training set of the original dataset. Meanwhile, the test set of the meta-dataset contains client model parameters trained on samples from the test set of the original dataset. This train/test split reflects an assumption that the proxy data accessible by the adversary is different from the training data on IoT devices, but drawn from the same distribution. To generate synthetic non-IID client label distributions on which dummy clients are trained, we sample from a symmetric Dirichlet distribution, where the concentration parameter $\alpha$ governs the uniformity of the client label distribution. To capture a wide range of uniformities, we construct the meta-dataset using an equal number of client label distributions with each of the concentration parameters $\alpha \in \{0.1, 1, 10, 100, 1000\}$. Example client label distributions are shown in \autoref{fig:dirichlet_examples}. For each experiment, the training split of the meta-dataset contains 10,000 samples in total (2,000 per value of $\alpha$), and the test split contains 2,000 samples in total (400 per value of $\alpha$). We use a simplified model in which the adversary only performs the attack during the first round of FedAvg. In other words, the adversary intercepts the initial global model $w_0$ and the client models $w_1^k$. In constructing the meta-dataset, we reduce the client model parameters to 10 dimensions via PCA.

For each level of noise, we first measure the accuracy of a global model trained with FedAvg across 500 rounds, in which training samples are assigned to clients uniformly at random. In our tests, we use $K = 100$ clients, each receiving 500 training samples. We use client fraction of $C = 0.1$ and train each client for $E = 1$ local epoch before aggregation. For MNIST, the client updates use a batch size of 32 and a learning rate of $10^{-4}$; for CIFAR-10, the updates use a batch size of 32 and a learning rate of $10^{-3}$. To reduce the effects of random model initialization and client sampling, we perform eight trials for each dataset and level of noise.

Then, we train a label distribution predictor on the training set of the meta-dataset and assess its performance on the test set. The predictor uses an MLP architecture consisting of an input layer with 10 units, eight hidden layers with 1000 units each and ReLU activations, and an output layer with 10 units and a softmax activation to produce a valid probability distribution over the labels.

\begin{figure*}
	\centering
    \includegraphics[width=0.95\linewidth]{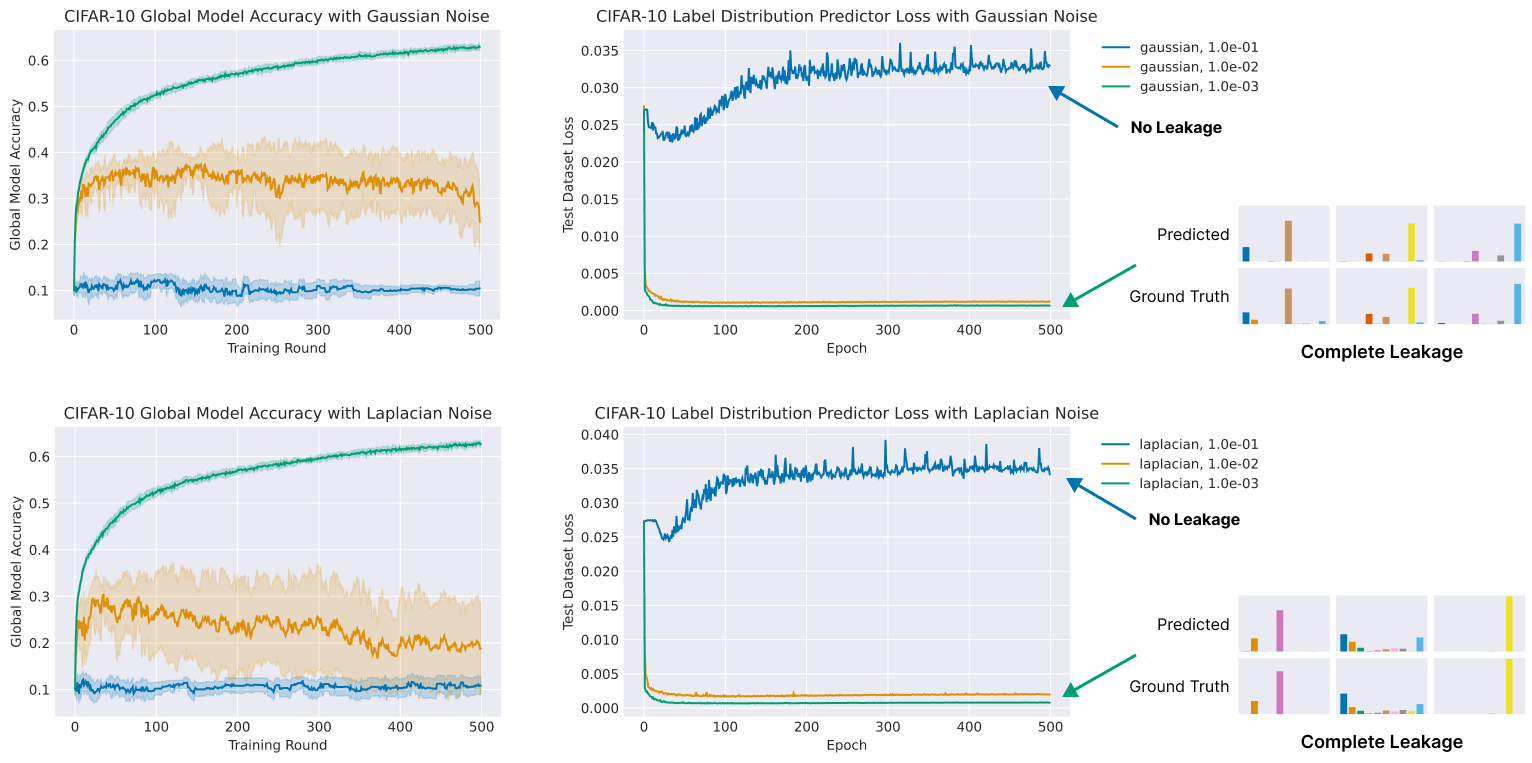}
    \caption{On the left is the test accuracy of the global model trained on the CIFAR-10 dataset with FedAvg; on the right is the test loss of the label distribution predictor trained on the meta-dataset. Adding Gaussian or Laplacian noise appears to be insufficient in protecting against data distribution leakage. In both cases, the scale of the noise required ($10^{-1}$) prevents the global model from learning.}
    \label{fig:cifar_label_distribution}
\end{figure*}

The results of the meta-dataset training are shown for the CIFAR-10 dataset in \autoref{fig:cifar_label_distribution}; experiments on MNIST showed almost identical results. We find that with $10^{-3}$-scale Gaussian and Laplacian noise, the global model is able to learn, but the adversary can completely uncover the client data distributions. Furthermore, even with $10^{-2}$-scale noise, when the model performance is notably compromised, we found almost no benefit in preventing privacy leakage. Consequently, our findings suggest that the adversarial strategy proposed in Algorithm \ref{alg:adversary} is capable of predicting client label distributions with high accuracy, and adding noise to client gradients is insufficient in protecting against the adversarial model.

%% file: sections/6_conclusion.tex
We have introduced an algorithm by which, in realistic settings, an adversary can predict the label distributions of federated clients. Our approach lays the groundwork for a general class of algorithms in which an adversary creates a meta-dataset which maps client model parameters to some property of their local data. Future work might investigate other approaches within this class of algorithms, such as predicting the presence of a particular type of sample, rather than a full label distribution. More research is also necessary to determine the effectiveness of the proposed adversarial technique on more complex datasets and with self-supervised training objectives. Finally, we have shown that injecting noise into client gradients is insufficient to protect privacy. Further work is needed to determine the effectiveness of other defensive strategies in limiting these forms of attacks.